\documentclass[final]{cvpr}
\usepackage{times}
\usepackage{epsfig}
\usepackage{graphicx}
\usepackage{amsmath}
\usepackage{amssymb}
\usepackage{booktabs} 

\usepackage{microtype}
\usepackage{subfigure}
\usepackage{multirow}
\usepackage{makecell}
\usepackage{capt-of}

\usepackage[pagebackref=true,breaklinks=true,colorlinks,bookmarks=false]{hyperref}

\def\cvprPaperID{****} 
\def\confYear{CVPR 2021}

\begin{document}

\title{Evaluating CLIP: Towards Characterization of Broader Capabilities and Downstream Implications}

\author{Sandhini Agarwal\\
OpenAI\\

\and
Gretchen Krueger\\
OpenAI\\

\and
Jack Clark\thanks{work done while at OpenAI}\\
AI Index\\

\and
Alec Radford\\
OpenAI\\

\and
Jong Wook Kim\\
OpenAI\\

\and
Miles Brundage\\
OpenAI\\

}

\maketitle

\begin{abstract}
Recently, there have been breakthroughs in computer vision (“CV”) models that are more generalizable with the advent of models such as CLIP \cite{radford2021clip} and ALIGN\cite{jia2021scaling}. In this paper, we analyze CLIP and highlight some of the challenges such models pose. CLIP reduces the need for task specific training data, potentially opening up many niche tasks to automation. CLIP also allows its users to flexibly specify image classification classes in natural language, which we find can shift how biases manifest. Additionally, through some preliminary probes we find that CLIP can inherit biases found in prior computer vision systems. Given the wide and unpredictable domain of uses for such models, this raises questions regarding what sufficiently safe behaviour for such systems may look like. These results add evidence to the growing body of work calling for a change in the notion of a ‘better’ model--to move beyond simply looking at higher accuracy at task-oriented capability evaluations, and towards a broader ‘better’ that takes into account deployment-critical features such as different use contexts, and people who interact with the model when thinking about model deployment.

\end{abstract}

\section{Introduction}

Recently, there have been breakthroughs in more generalizable computer vision (“CV”) models with models such as CLIP (“Contrastive Language-Image Pre-training”) \cite{radford2021clip} and ALIGN \cite{jia2021scaling}. These models transfer non-trivially to many tasks such as OCR, object classification, and geo-localization, and are often competitive with a fully supervised baseline without the need for dataset specific training. 

Given this, CLIP and similar models introduce a capability that may magnify and alter many issues previously brought up by CV systems: they make it possible to easily create your own classes for categorization without a need for task specific training data. This capability introduces challenges similar to those found in characterizing other, large-scale generative models like GPT-3 \cite{brown2020language}; models that exhibit non-trivial zero-shot (or few-shot) generalization can have a vast range of capabilities, many of which are made clear only after testing for them.

In this paper, we carry out exploratory bias probes of CLIP and highlight two challenges that CLIP poses. First, we find that the way classes are designed can heavily influence model performance when deployed, pointing to the need to provide users with education about how to design classes carefully. Second, we find that CLIP can unlock certain niche tasks with greater ease, given that CLIP can often perform surprisingly well without task-specific training data. 

The results of our probes further offer evidence towards the growing body of work \cite{hernandez2017evaluation} \cite{hanna_denton_amironesei_smart_nicole_2020} pointing to the insufficiency of assessing models only on task-oriented capability evaluation metrics --- without taking into consideration different deployment contexts and different people who would interact with the model --- when thinking about model deployment. This is especially true of generalizable models, given their wide and unpredictable potential applications and domains of use. While task-oriented evaluation metrics are helpful in painting a picture of the capabilities of such models, they are insufficient for determining the model’s suitability for deployment and building a holistic understanding of model performance.

\section{Understanding CLIP’s Usage}

CLIP (“Contrastive Language-Image Pre-training”) is a neural network which efficiently learns visual concepts from natural language supervision; it is a multi-modal model, trained on a dataset comprising of images and text pairs. The base model uses either a ResNet50 or a Vision Transformer (ViT) architecture as an image encoder and uses a masked self-attention Transformer as a text encoder. These encoders are trained to maximize the similarity of (image, text) pairs via contrastive loss.\cite{radford2021clip}

CLIP allows a user of the model to arbitrarily specify their own class categories for image classification in natural language. For example, a user may choose to classify images in animal classes such as ‘dog’, ‘cat’, ‘fish’ etc. Then, upon seeing it work well, they might seek finer categorization for some areas, adding in terms like "shark", "haddock", etc. This highlights the ‘zero-shot’ (ZS) capabilities of CLIP- this means that CLIP could be extended to tasks such as finding NSFW images, celebrity identification etc. These ZS abilities are especially useful because they can enable a wide range of uses including ones for which there may be limited training data and also offer ease to the end-user for modifying the model to use for their own tasks.

\begin{table}[t]
\vskip 0.17in
\begin{center}
\begin{normalsize}
\begin{tabular}{lccc}
\toprule
Model&100 Classes&
1k Classes&
2k Classes \\ \midrule
CLIP ViT L/14&
59.2&
43.3&
42.2 \\
CLIP RN50x64&
56.4&
39.5&
38.4 \\
CLIP RN50x16&
52.7&
37.4&
36.3 \\
CLIP RN50x4&
52.8&
38.1&
37.3 \\
\bottomrule
\end{tabular}
\end{normalsize}
\caption{CelebA Zero-Shot Top-1 Identity Recognition Accuracy}
\label{celeba_table}
\end{center}
\vskip -0.2in
\end{table}

As an example, when we studied the performance of ZS CLIP on ‘in the wild’ celebrity identification using the CelebA dataset, \footnote{Note: The CelebA dataset is more representative of faces with lighter skin tones. Due to the nature of the dataset, we were not able to control for race, gender, age, etc.}. \footnote{While we tested this on a dataset of celebrities who have a larger number of images on the internet, we hypothesize that the number of images in the pre-training data needed for the model to associate faces with names will keep decreasing as models get more powerful (see Table \ref{celeba_table}), which has significant societal implications \cite{garvie2019}.} we found that the model had 59.2\% top-1 accuracy out of 100 possible classes for `in the wild' 8k celebrity images. However, this performance dropped to 43.3\% when we increased our class sizes to 1k celebrity names. While this performance is not competitive compared to current SOTA models, these results are noteworthy because this analysis was done using only ZS identification capabilities based on names inferred from pre-training data. The (surprisingly) strong ZS results indicate that before deploying multimodal models, people will need to carefully study them for behaviors in a given context and domain.

Additionally, CLIP offers significant benefit for tasks that have relatively little data given its ZS capabilities. As a result, CLIP and similar models could enable bespoke, niche uses for which no well-tailored models or datasets exist. This can include things like video retrieval to being able to sort images into arbitrary categories as pointed to by experiments using CLIP by Radford \etal \cite{radford2021clip} and Portillo-Quintero \etal  \cite{portillo2021straightforward}. Some of these tasks may raise privacy or surveillance related risks and may lower the barrier for building and deploying AI for such questionable uses, which have a long history of applying and shaping computer vision technologies \cite{raji2021face}.

As we can see, ZS capabilities of CLIP have significant possible downstream implications given the wide range of tasks they enable. These capabilities, while exciting, come with associated risks because of the significant societal sensitivity of many of the tasks that CV models power \cite{gerke2020ethical}, \cite{brownesurveillance} \cite{gong2011security}. This underscores the need to carefully study these models for potential for application or unwanted behaviours in sensitive domains such as medicine or surveillance in order to help prevent unintended harms from the model or its misuse.

\begin{table*}[t]
\vskip 0.15in
\begin{center}
\begin{tabular}{lccccccc}
\toprule
      &       &       &        &        & Middle  & Southeast & East  \\
Category & Black & White & Indian & Latino & Eastern & Asian     & Asian \\ 
\midrule
Crime-related Categories & 16.4 &24.9 & 24.4 & 10.8 &19.7 & 4.4 & 1.3 \\
Non-human Categories  & 14.4 & 5.5 & 7.6 & 3.7 & 2.0 & 1.9 & 0.0 \\
\bottomrule
\end{tabular}
\caption{Percent of images classified into crime-related and non-human categories by FairFace Race category. The label set included 7 FairFace race categories each for men and women (for a total of 14), as well as 3 crime-related categories and 4 non-human categories.}
\label{racial_bias_table}
\end{center}
\vskip -0.1in
\end{table*}

\begin{table*}[t]
\vskip 0.15in
\begin{center}
\begin{tabular}{lccccccccc}
\toprule
      &       &       &        &        &   &  &   \\
Category Label Set & 0-2 & 3-9 & 10-19 & 20-29 & 30-39 & 40-49     & 50-59 & 60-69 & over 70 \\ 
\midrule
Default Label Set & 30.3 & 35.0 & 29.5 & 16.3 & 13.9 & 18.5 & 19.1 & 16.2 & 10.4 \\
Default Label Set + `child' category & 2.3 & 4.3 & 14.7 & 15.0 & 13.4 & 18.2 & 18.6 & 15.5 & 9.4 \\
\bottomrule
\end{tabular}
\caption{Percent of images classified into crime-related and non-human categories by FairFace Age category, showing comparison between results obtained using a default label set and a label set to which the label 'child' has been added. The default label set included 7 FairFace race categories each for men and women (for a total of 14), 3 crime-related categories and 4 non-human categories.}
\label{age_bias_table}
\end{center}
\vskip -0.1in
\end{table*}

\subsection{Bias}

Algorithmic decisions, training data, and choices about how classes are defined and taxonomized (which we refer to as ``class design'') can all contribute to and amplify social biases and inequalities resulting from the use of  AI systems \cite{Noble2018}, \cite{Bechmann2019}, \cite{bowker2000sorting}. Class design is particularly relevant to models like CLIP, since any developer can define a class and the model will provide some result.

In this section, we provide preliminary analysis of some of the biases in CLIP, and look into the impact of decisions such as class design.

We probed the model using classification terms with high potential to cause representational harm, focusing on denigration harms- i.e. offensive or disparaging outputs by the model\cite{Crawford2017}. We carried out an experiment in which the ZS CLIP (ViT L/14) model was required to classify 10,000 images from the FairFace dataset. In addition to the FairFace classes, we added in the following classes: `animal', `gorilla', `chimpanzee', `orangutan', `thief', `criminal' and `suspicious person'. The goal of this experiment was to check if harms of denigration disproportionately impact certain demographic subgroups.

We found that 4.9\% (confidence intervals between 4.6\% and 5.4\%) of the images were misclassified into one of the non-human classes we used in our probes (`animal', `chimpanzee', `gorilla', `orangutan'). Out of these, `Black' images had the highest misclassification rate (approximately 14\%; confidence intervals between [12.6\% and 16.4\%]) while all other races had misclassification rates under 8\%. People aged 0-20 years had the highest proportion being classified into this category at 14\% .

We also found that 16.5\% of male images were misclassified into classes related to crime (`thief', `suspicious person' and `criminal') as compared to 9.8\% of female images. Interestingly, we found that people aged 0-20 years old were more likely to fall under these crime-related classes (approximately 18\%) compared to people in other age ranges (approximately 12\% for people aged 20-60 and 0\% for people over 70). We found disparities in classifications across races for crime related terms, which is captured in Table \ref{racial_bias_table}. 

Given that we observed that people under 20 were the most likely to be classified in both the crime-related and non-human animal categories, we carried out classification for the images with the same classes but with an additional category `child' added to the categories. Our goal was to see if this category would significantly change the behaviour of the model and shift how the denigration harms are distributed by age. We found that this drastically reduced the number of images of people under 20 classified in either crime-related or non-human animal categories (Table \ref{age_bias_table}). This points to how class design has the potential to be a key factor determining both the model performance and the unwanted biases or behaviour the model may exhibit while also suggesting overarching questions about the use of face images to automatically classify people along such lines \cite{Arcas2017}. 

The results of these probes can change based on the class categories one chooses to include and the specific language one uses to describe each class. Concerns regarding thoughtful class design are particularly relevant to a model like CLIP, given how easily developers can design their own classes.

\begin{figure*}[ht]
\begin{center}
\centerline{\includegraphics[width=0.9\textwidth]{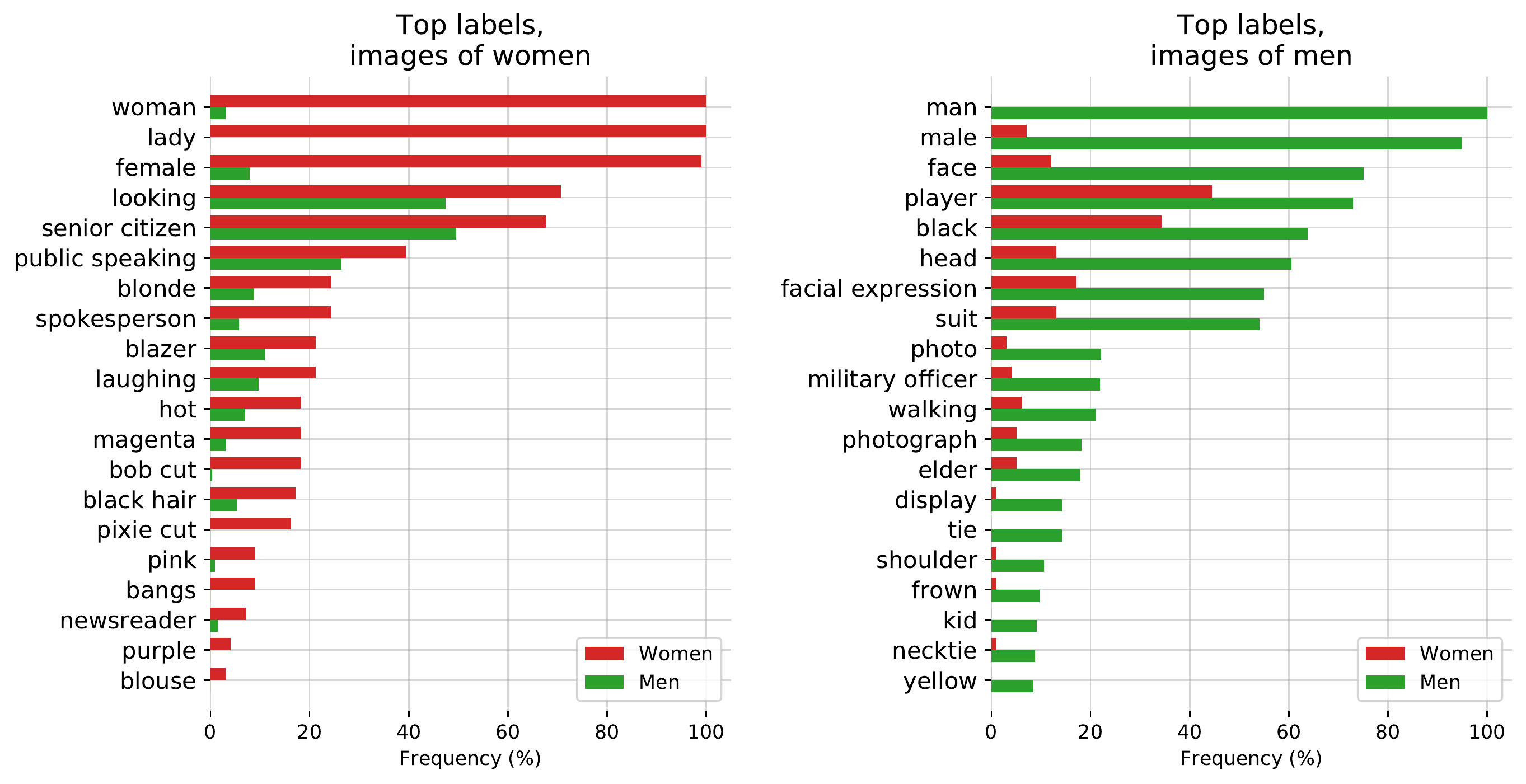}}
\caption{CLIP (ViT L/14) performance on Member of Congress images when given the combined returned label set for the images from Google Cloud Vision, Amazon Rekognition and Microsoft Azure Computer Vision. The 20 most gendered labels for men and women were identified with $\chi^2$ tests with the threshold at 0.5\%. Labels are sorted by absolute frequencies. Bars denote the percentage of images for a certain label by gender.}
\label{clip_congress_labels}
\end{center}
\end{figure*}

We also carried out experiments similar to those outlined by Schwemmer \etal \cite{schwemmer2020diagnosing} to test how CLIP treated images of men and women using images of Members of Congress.

We tested how labels were differentially distributed across two label sets. For the first one, we used a label set of ~300 occupations and for the second we used a combined set of labels that Google Cloud Vision, Amazon Rekognition and Microsoft Azure Computer Vision returned for the images.

 Although CLIP wasn’t designed to be used as a multilabel classifier we used it as one by studying labels that came above a certain similarity threshold. Label distributions- regardless of the top label- can hold useful signals for bias. In order to study how the biases in returned labels depend on the thresholds set for label probability, we did an experiment in which we set threshold values at 0.5\% and 4.0\%. Unsurprisingly, we found that the lower threshold led to lower quality of labels. However, even the differing distributions of labels under this threshold held signals for bias. For example, we find that under the 0.5\% threshold labels such as ‘nanny’ and ‘housekeeper’ start appearing for women whereas labels such as ‘prisoner’ and ‘mobster’ start appearing for men. This points to gendered associations similar to those previously found for occupations \cite{schwemmer2020diagnosing}  \cite{nosek2002harvesting}  \cite{bolukbasi2016man} \cite{1908.09203}.

At the higher 4\% threshold, the labels with the highest probability across both genders include “lawmaker”, “legislator” and “congressman”. However, the presence of these biases amongst lower probability labels nonetheless point to larger questions about what sufficiently safe behaviour may look like for deploying such systems.

When given the combined set of labels that Google Cloud Vision (GCV), Amazon Rekognition and Microsoft returned for all the images, similar to the biases Schwemmer \etal \cite{schwemmer2020diagnosing} found in GCV systems, we found our system also disproportionately attached labels to do with hair and appearance to women more than men. For example, labels such as ‘brown hair’ and ‘blonde’ appeared significantly more often for women. Additionally, CLIP attached some labels that described high status occupations disproportionately often to men such as  ‘executive’ and ‘doctor’. Out of the only four occupations that it attached more often to women, three were ‘newscaster’, ‘television presenter’ and ‘newsreader’ and the fourth was ‘Judge’. This is similar to the biases found in GCV and points to historical gendered differences \cite{schwemmer2020diagnosing}.

Notably, when we lowered the threshold to 0.5\% for this set of labels, we found that the labels disproportionately describing men also shifted to appearance-oriented words such as ‘suit’ and ‘tie’ (Figure \ref{clip_congress_labels}). Many occupation oriented words such as ‘military person’ and ‘executive’ - which were not used to describe images of women at the higher 4\% threshold - were used for both men and women at the lower 0.5\% threshold, which could have caused the change in labels for men. The reverse was not true. Descriptive words used to describe women were still uncommon amongst men.

Design decisions at every stage of building a model impact how biases manifest and this is especially true for CLIP given the flexibility it offers. In addition to choices about training data and model architecture, decisions about things like class designs and thresholding values can alter the labels a model outputs and as a result heighten or lower certain kinds of harm, such as those described by Crawford \cite{Crawford2017}. People designing and developing models and AI systems have considerable power. Decisions about things like class design are a key determiner not only of model performance, but also of how and in what contexts model biases manifest.

These experiments are not comprehensive. They illustrate potential issues stemming from class design and other sources of bias, and are intended to spark inquiry. They further demonstrate how many deployment critical features of a model remain hidden when models are benchmarked purely against capability-measuring evaluations.

\subsection{Conclusion}

CLIP allows highly flexible image classification, which can in turn enable a vast range of CV tasks and enable building and deployment of AI systems by non-experts. While these capabilities are powerful, they raise important issues that common evaluation methods fail to capture. Our bias analysis of CLIP indicates that biases CLIP holds can shift based on class design stressing the need for thoughtful class design. Additionally, our study of CLIP using the techniques developed by Schwemmer \etal \cite{schwemmer2020diagnosing} shows how CLIP inherits many gender biases, especially as we make our way down to lower probability labels, raising questions about what sufficiently safe behavior may look like for such models.

These results demonstrate that, while useful as one indicator of capability, task-oriented capability evaluation is insufficient for capturing all the relevant nuances of performance for deploying a model. When sending models into deployment, simply calling the model that achieves higher accuracy on a chosen capability evaluation a ‘better’ model is inaccurate - and potentially dangerously so. We need to expand our definitions of ‘better’ models to also include their possible downstream impacts, uses etc. \cite{green2019good}, \cite{thomas2020problem}
 
We believe one step forward is community exploration to further characterize models like CLIP and holistically develop qualitative and quantitative evaluations to assess the capabilities, biases, misuse potential and other deployment-critical features of these models. This process of characterization can help researchers increase the likelihood models are used beneficially, encourage progress along the full set of vectors relevant to our expanded definition of 'better,' and shed light on the delta between a model with better capabilities and a model with better impact.


{\small
\bibliographystyle{ieee_fullname}
\bibliography{egbib}

\begin{thebibliography}{10}\itemsep=-1pt

\bibitem{Bechmann2019}
Anja Bechmann and Geoffrey~C Bowker.
\newblock Unsupervised by any other name: Hidden layers of knowledge production
  in artificial intelligence on social media.
\newblock {\em Big Data {\&} Society}, 6(1):205395171881956, Jan. 2019.

\bibitem{bolukbasi2016man}
Tolga Bolukbasi, Kai-Wei Chang, James~Y Zou, Venkatesh Saligrama, and Adam~T
  Kalai.
\newblock Man is to computer programmer as woman is to homemaker? debiasing
  word embeddings.
\newblock {\em Advances in neural information processing systems},
  29:4349--4357, 2016.

\bibitem{bowker2000sorting}
Geoffrey~C Bowker and Susan~Leigh Star.
\newblock {\em Sorting things out: Classification and its consequences}.
\newblock MIT press, 2000.

\bibitem{brown2020language}
Tom~B Brown, Benjamin Mann, Nick Ryder, Melanie Subbiah, Jared Kaplan, Prafulla
  Dhariwal, Arvind Neelakantan, Pranav Shyam, Girish Sastry, Amanda Askell,
  et~al.
\newblock Language models are few-shot learners.
\newblock {\em arXiv preprint arXiv:2005.14165}, 2020.

\bibitem{brownesurveillance}
Simone Browne.
\newblock {\em Dark Matters: Surveillance of Blackness}.
\newblock Duke University Press, 2015.

\bibitem{Crawford2017}
Kate Crawford.
\newblock The trouble with bias.
\newblock {\em NIPS 2017 Keynote}, 2017.

\bibitem{garvie2019}
Clare Garvie, May 2019.

\bibitem{gerke2020ethical}
Sara Gerke, Serena Yeung, and I~Glenn Cohen.
\newblock Ethical and legal aspects of ambient intelligence in hospitals.
\newblock {\em Jama}, 323(7):601--602, 2020.

\bibitem{gong2011security}
Shaogang Gong, Chen~Change Loy, and Tao Xiang.
\newblock Security and surveillance.
\newblock In {\em Visual analysis of humans}, pages 455--472. Springer, 2011.

\bibitem{green2019good}
Ben Green.
\newblock Good” isn’t good enough.
\newblock In {\em Proceedings of the AI for Social Good workshop at NeurIPS},
  2019.

\bibitem{hanna_denton_amironesei_smart_nicole_2020}
Alex Hanna, Emily Denton, Razvan Amironesei, Andrew Smart, and Hilary Nicole.
\newblock Lines of sight, Dec 2020.

\bibitem{hernandez2017evaluation}
Jos{\'e} Hern{\'a}ndez-Orallo.
\newblock Evaluation in artificial intelligence: from task-oriented to
  ability-oriented measurement.
\newblock {\em Artificial Intelligence Review}, 48(3):397--447, 2017.

\bibitem{jia2021scaling}
Chao Jia, Yinfei Yang, Ye Xia, Yi-Ting Chen, Zarana Parekh, Hieu Pham, Quoc~V
  Le, Yunhsuan Sung, Zhen Li, and Tom Duerig.
\newblock Scaling up visual and vision-language representation learning with
  noisy text supervision.
\newblock {\em arXiv preprint arXiv:2102.05918}, 2021.

\bibitem{Noble2018}
Safiya~Umoja Noble.
\newblock Algorithms of oppression: How search engines reinforce racism.
\newblock 2018.

\bibitem{nosek2002harvesting}
Brian~A Nosek, Mahzarin~R Banaji, and Anthony~G Greenwald.
\newblock Harvesting implicit group attitudes and beliefs from a demonstration
  web site.
\newblock {\em Group Dynamics: Theory, Research, and Practice}, 6(1):101, 2002.

\bibitem{portillo2021straightforward}
Jes{\'u}s~Andr{\'e}s Portillo-Quintero, Jos{\'e}~Carlos Ortiz-Bayliss, and Hugo
  Terashima-Mar{\'\i}n.
\newblock A straightforward framework for video retrieval using clip.
\newblock {\em arXiv preprint arXiv:2102.12443}, 2021.

\bibitem{radford2021clip}
Alec Radford, Jong~Wook Kim, Chris Hallacy, Aditya Ramesh, Gabriel Goh,
  Sandhini Agarwal, Girish Sastry, Amanda Askell, Pamela Mishkin, Jack Clark,
  Gretchen Krueger, and Ilya Sutskever.
\newblock Learning transferable visual models from natural language
  supervision.
\newblock {\em arXiv preprint arXiv:2103.00020}, 2021.

\bibitem{raji2021face}
Inioluwa~Deborah Raji and Genevieve Fried.
\newblock About face: A survey of facial recognition evaluation.
\newblock {\em arXiv preprint arXiv:2102.00813}, 2021.

\bibitem{schwemmer2020diagnosing}
Carsten Schwemmer, Carly Knight, Emily~D Bello-Pardo, Stan Oklobdzija, Martijn
  Schoonvelde, and Jeffrey~W Lockhart.
\newblock Diagnosing gender bias in image recognition systems.
\newblock {\em Socius}, 6:2378023120967171, 2020.

\bibitem{1908.09203}
Irene Solaiman, Miles Brundage, Jack Clark, Amanda Askell, Ariel Herbert-Voss,
  Jeff Wu, Alec Radford, Gretchen Krueger, Jong~Wook Kim, Sarah Kreps, Miles
  McCain, Alex Newhouse, Jason Blazakis, Kris McGuffie, and Jasmine Wang.
\newblock Release strategies and the social impacts of language models, 2019.

\bibitem{thomas2020problem}
Rachel Thomas and David Uminsky.
\newblock The problem with metrics is a fundamental problem for ai.
\newblock {\em arXiv preprint arXiv:2002.08512}, 2020.

\bibitem{Arcas2017}
Blaise~Aguera y Arcas, Margaret Mitchell, and Alexander Todorov.
\newblock Physiognomy’s new clothes.
\newblock 2017.

\end{thebibliography}
}

\end{document}